\documentclass[twoside]{article}

\usepackage[accepted]{aistats2020}
\usepackage[utf8]{inputenc} 
\usepackage[T1]{fontenc}    
\usepackage{hyperref}       
\usepackage{url}            
\usepackage{booktabs}       
\usepackage{amsfonts}       
\usepackage{nicefrac}       
\usepackage{microtype}      
\usepackage{algorithm}
\usepackage{algorithmic}
\usepackage{xcolor}

\usepackage{microtype}
\usepackage{graphicx}
\usepackage{subfigure}
\usepackage{booktabs} 
\usepackage{amsmath}
\usepackage{amssymb}
\usepackage{amsthm}
\usepackage{bbm}

\newtheorem{theorem}{Theorem}

\newtheorem{lemma}[theorem]{Lemma}

\DeclareMathOperator*{\argmax}{argmax}

\begin{document}

%

%

\twocolumn[

\aistatstitle{Thompson Sampling for Linearly Constrained Bandits}

\aistatsauthor{ Vidit Saxena \And Joseph E. Gonzalez \And  Joakim Jald\'{e}n }

\aistatsaddress{ KTH Royal Institute of Technology\\Stockholm, Sweden\\vidits@kth.se \And  University of California\\Berkeley\\jegonzal@berkeley.edu \And KTH Royal Institute of Technology\\Stockholm, Sweden\\jalden@kth.se } ]

\begin{abstract}
We address multi-armed bandits (MAB) where the objective is to maximize the cumulative reward under a probabilistic linear constraint. For a few real-world instances of this problem, constrained extensions of the well-known Thompson Sampling (TS) heuristic have recently been proposed. However, finite-time analysis of constrained TS is challenging; as a result, only $O\big(\sqrt{T}\big)$ bounds on the cumulative reward loss (i.e., the \emph{regret}) are available. In this paper, we describe LinConTS, a TS-based algorithm for bandits that place a linear constraint on the probability of earning a reward in every round. We show that for LinConTS, the regret as well as the cumulative constraint violations are upper bounded by $O(\log T)$ for the suboptimal arms. We develop a proof technique that relies on careful analysis of the dual problem and combine it with recent theoretical work on unconstrained TS. Through numerical experiments on two real-world datasets, we demonstrate that LinConTS outperforms an asymptotically optimal upper confidence bound (UCB) scheme in terms of simultaneously minimizing the regret and the violation.

\end{abstract}

\section{Introduction}
Multi-armed bandits (MAB) are a systematic way of modeling sequential decision problems. In MAB problems, an agent plays a sequence of arms aimed at optimizing some cumulative objective over $T$ dicrete time intervals (``rounds''). The agent strives to achieve this goal by sequentially \emph{exploring} the available arms and \emph{exploiting} historical rewards from previously selected arms. MABs have been successfully applied to problems in dynamic pricing, online procurement, and digital advertising, where the goal is to minimize the \emph{regret} (i.e., the cumulative reward loss) over a finite time horizon~\cite{russo2018tutorial}.

Despite the popularity of MABs, the unconstrained reward maximization goal does not extend to several commonly encountered sequential decision problems. In this paper, we consider a constrained MAB problem where the objective function is subject to a specific type of linear constraint, namely, that the probability of receiving a reward in any round exceeds a fixed, pre-defined threshold. A few recently studied applications of this type of constraint are listed below:

\textbf{Weblink selection:} Several online content providers earn revenue by advertising affiliate weblinks on their page. In this context, the goal is to select a subset of links from a pool of available links, which collectively maximize the provider's ad revenue. At the same time, the provider typically wants to avoid displaying some high-revenue links that have a low \emph{click-through rate} (CTR), since such links may be perceived as clutter and drive down user satisfaction. In~\cite{chen2018beyond}, this problem is formulated as a constrained MAB where the probability of clicking on one of the displayed weblinks must exceed a minimum CTR threshold.

\textbf{Wireless Rate selection:} Wireless communication networks strive to optimize the data transmission rate for packetized information bits. Aggressive data transmission rates carry more per-packet information but also suffer from more frequent packet failures compared to conservative rates. For latency-sensitive applications such as video streaming and online gaming, the goal of wireless rate selection is to maximize the average data throughput while simultaneously maintaining a minimum packet success frequency. Recently,~\cite{saxena2019constrained} proposed a constrained MAB approach for this rate selection problem that was shown to outperform state-of-the-art techniques.

Thompson sampling (TS) is an efficient Bayesian heuristic for MAB optimization~\cite{thompson1933likelihood, russo2018tutorial}. TS operates by conditioning the probability of selecting an optimal arm on the historical rewards observed in previous rounds. The regret for TS has been shown to scale as $O(\log T)$ in~\cite{kaufmann2012thompson, agrawal2013further}, where the constant factors depend on the distribution of the underlying problem parameters. These bounds asymptotically achieve the optimal regret established in~\cite{lai1985asymptotically}. Further,~\cite{ferreira2018online} proposed TS-based algorithm for a stochastic packing problem commonly referred to as \emph{Bandits with Knapsacks (BwK)}, for which the distribution-independent regret upper bounds were shown to scale as $O(\sqrt{T})$. For BwK, $O(\log T)$ distribution-dependent lower bounds were developed in~\cite{flajolet2015logarithmic}. 

In this paper, we propose \emph{LinConTS}, a TS-based algorithm for the specific type of linear constraint described above. Our algorithm encodes a probabilistic arm selection policy, where the arm selection weights are obtained by solving a linear program (LP) subroutine in every round. We denote the arms supported by the stationary optimal policy as \emph{optimal arms} and the rest as \emph{suboptimal arms}. We show that for the rounds where the suboptimal arms are played, the LinConTS regret is upper bounded by $O(\log T)$. Further, we show that for these arms, the \emph{violation} metric, i.e., the cumulative number of constraint violations, is also upper bounded by $O(\log T)$. For the rounds where the optimal arms are played, the regret and violation metrics scale only as fast as $O(\sqrt{T})$\footnote{The original version of this paper cited prior work to claim that the regret and violation contribution of optimal arms also scales as $O(\log T)$. In fact, prior work only establishes $O(\sqrt{T})$ bounds, which has been updated in this version of our paper.}. As such, our theoretical results constitute a significant tightening of the regret and violation bounds available in the prior literature.

\emph{Proof technique:} The recipe for MABs with linearly constrained objectives is to invoke a LP subroutine in each round. Subsequently, the central problem in analyzing constrained MABs arises from the fact that the arm parameters cannot be compared directly; instead, they interact with each other through the LP. To handle this issue, our proof technique relies on careful examination of the dual LP problem. In particular, to bound the probability of selecting a particular arm, we analyze stochastic perturbations of the convex polytope enclosing the LP feasible region. We use these perturbation results in conjunction with theoretical analysis of TS developed in~\cite{agrawal2013further}.

\section{Problem Formulation}
\label{sec:problem}
Consider a MAB with $N$ arms, where at each time step $t=1,2,\dots,T$, one of the arms must be played. Playing the arm $i(t)\in\{1,\dots,N\}$ results in a Bernoulli-distributed \emph{reward event} with mean $\mu_{i(t)}$. Reward events are independent across the arms and and across successive plays of an arm. In case of a successful reward event, a reward value $r_{i(t)}\in (0, 1]$ is immediately collected by the player. We assume that the $\mu_i$s are \emph{a priori} unknown while the $r_i$s are deterministic and known in advance. In the context of weblink selection, the arms $(\mu_i,r_i)$ correspond to the pool of available weblinks. For a selected weblink that is displayed on a target webpage, the reward event denotes a Bernoulli-distributed user click with an unknown mean CTR $\mu_i$. Subsequently, the reward value $r_i$ captures the revenue generated when a user clicks on the displayed link. If a user clicks on the displayed link, the reward event is deemed to be true and the corresponding reward value is collected. On the other hand if a user fails to click on the link, the reward event is false and no reward is collected in that round.

The problem objective is to maximize the cumulative expected reward value, $\sum_{t=1}^T \mu_{i(t)}r_{i(t)}$ subject to the constraint $(1/T)\sum_{t=1}^T\mu_{i(t)}\geq\eta$ where $\eta\in[0,1]$. The constraint, which ensures that at least a fraction $\eta$ of rounds result in a reward event, must be satisfied with a high probability. This problem can be formulated using the LP relaxation
\begin{align}
    LP(\boldsymbol\mu):\,\text{maximize}\,&\sum_i x_i\mu_ir_i \nonumber \\
    \text{subject to}\,&\sum_i x_i\mu_i \geq \eta, \nonumber \\
    &\sum_i x_i
    =1, \nonumber \\
    & x_i \geq 0,\,i=1,\dots,N.
    \label{eq:linear_prog}
\end{align}
where $x_i$ denotes probability of selecting arm $i$ and therefore $\boldsymbol{x}=[x_1,\dots,x_N]$ is the \emph{probabilistic arm selection vector}. The arms assigned probability mass $x_i>0$ by the LP solution constitute the set of \emph{optimal} arms that are played with a non-zero probability.

\emph{Stationary optimal policy:} If the $\mu_i$s were known in advance, the stationary optimal policy is found by solving $LP(\boldsymbol\mu)$. We denote the solution of this LP with $\boldsymbol{x}^*=[x_1^*,\dots,x_N^*]$, which is the optimal stationary probabilistic arm selection vector. The expected reward of the stationary optimal policy is therefore $r^*=\sum_ix_i^*\mu_ir_i$. The arms assigned a probability mass $x_i^*>0$ by the optimal policy constitute the set of stationary \emph{optimal arms} and the rest of the arms are stationary \emph{suboptimal arms}. 

\emph{Assumptions:} We assume that the arm with the highest expected reward value, $i_\text{max}=\argmax_i\mu_ir_i$, does not also satisfy the constraint, i.e., $\mu_{i_\text{max}}<\eta$ (otherwise the problem reduces to an unconstrained MAB problem). Consequently, to simultaneously satisfy the reward maximization and the constraint satisfaction objectives, a mixture of high-reward-value and high-reward-event-probability arms is required. Further, we assume that a strictly feasible solution exists for $LP(\boldsymbol\mu)$ and that there are no degenerate cases (i.e., the stationary optimal policy is a unique mixture of the arms).

\emph{Regret and violation:} The performance of any MAB algorithm is typically measured in terms of its regret, $\mathcal{R}(T)$, which is the cumulative reward loss compared to selecting the reward-maximizing arm in hindsight. However, in a constrained MAB setting, the optimal arms balance the need for reward maximization with constraint fulfilment. As a consequence, it is possible for a sub-optimal policy to exceed the cumulative reward achieved by the optimal policy, e.g., by frequently selecting a high-reward arm that instead violates the constraint. Therefore, we additionally define a \emph{violation} metric $\mathcal{V}(T)$ to measure the cumulative constraint violations until time $T$. For the estimated reward event probabilities in the $t^\text{th}$ round, $\widetilde{\boldsymbol\mu}_t=[\widetilde{\mu}_{1,t},\dots,\widetilde{\mu}_{N,t}]$, let the instantaneous, probabilistic arm selection vector is $x_t=[x_{1,t},\dots,x_{N,t}]$. Then the expected violation and regret until time $T$ are given by
\begin{align}
    \mathbb{E}\big[\mathcal{R}(T)\big]&=\mathbb{E}\bigg[\bigg[Tr^* - \sum_{t=1}^T\sum_i x_{i,t}\mu_ir_i\bigg]_+\bigg] \\
                  &= \bigg[\sum_i\Delta_i\mathbb{E}\big[k_i(T+1)\big]\bigg]_+
    \label{eq:regret},\\
    \mathbb{E}\big[\mathcal{V}(T)]&=\mathbb{E}\bigg[\bigg[T\eta - \sum_{t=1}^T\sum_ix_{i,t}\mu_i\bigg]_+\bigg] \\
                  &=\bigg[\sum_i(\eta - \mu_i)\mathbb{E}\big[k_i(T+1)\big]\bigg]_+,
    \label{eq:violation}
\end{align}
where $k_i(t)$ denotes the number of times that the arm $i$ is played until time $t-1$, $\Delta_i = r^*-\mu_ir_i$ is the expected loss in reward for arm $i$, and $[x]_+=\max\{x,0\}$. Intuitively, the reward metric measures the amount of reward lost due to selecting suboptimal arms, or due to selecting optimal arms with a frequency different from the stationary optimal policy. In order to minimize regret, a policy could simply choose to search for the reward-maximizing arm without optimizing for the constraint. To account for this behaviour, the violation metric keeps track of the constraint violations of the policy. Both regret and violation metrics must be simultaneously optimized by any useful policy. In the next sections, we develop finite-time upper bounds on regret and violation metrics for LinConTS.

\section{Algorithm and Finite-Time Analysis}
\label{sec:linconts}
\subsection{LinConTS}

\begin{algorithm}[tb]
   \caption{LinConTS}
   \label{alg:ts_linprog}
\begin{algorithmic}[1]
   \STATE {\bfseries Input:} Reward Values $r_{\{1,\dots,N\}}$, Constraint $\eta$
   \STATE \textbf{Initialize:} $\alpha_{\{1,\dots,N\},0}=1, \beta_{\{1\dots N\},0}=1$.
   \FOR{ Time index $t=1$ {\bfseries to} $T$ }
   \IF{ $t<N$ }
   \STATE $i(t) = t$
   \ELSE
   \STATE Sample $\widetilde{\mu}_{i,t}\sim \text{Beta}(\alpha_{i,t-1}, \beta_{i,t-1})$ for each arm $i=1,\dots,N$.
   \STATE Solve, if feasible, the linear program:
   \begin{align*}
  LP(\widetilde{\boldsymbol\mu}_t):\,\text{maximize}\,\sum_i x_{i,t}\widetilde{\mu}_{i,t}r_i \qquad\qquad\qquad\qquad
  \end{align*}
  \begin{equation}
  \qquad\text{ subject to } \begin{cases}
     \sum_i x_{i,t}\widetilde{\mu}_{i,t} \geq \eta\\
    \sum_ix_{i,t} = 1\\
    x_{i,t} \geq 0 \forall\,i\in\{1,\dots,N\}
  \end{cases},
  \label{eq:linconts}
\end{equation}
   \IF{ a (feasible) optimal solution exists } 
    \STATE Sample $i(t)~\sim [x_{1,t},\dots,x_{N,t}]$
   \ELSE  
    \STATE Sample $i(t)$ uniformly from $\{1,\dots,N\}$.
   \ENDIF
   \ENDIF
   \STATE \textbf{Observe:} Reward event $c_{i(t),t}\in\{0,1\}$.
   \STATE \textbf{Update:} \\
             $\qquad \alpha_{i(t),t} = \alpha_{i(t),t-1} + c_{i(t),t}$ \\
             $\qquad \beta_{i(t),t} = \beta_{i(t),t-1} + (1-c_{i(t),t})$.
   \ENDFOR
\end{algorithmic}
\label{alg:linconts}
\end{algorithm}

TS assigns prior distributions over the unknown MAB parameters. Subsequently for the played arm in evey round, the arm posterior is updated using the observed reward. For the problem studied in this paper, the unknown parameters are the reward event means $\mu_1,\dots,\mu_N$. Since these parameters are Bernoulli-distributed, a suitable choice of prior is the Beta distribution~\cite{russo2018tutorial}. The LinConTS algorithm is described in Alg.~\ref{alg:linconts}. At the beginning of the experiment, the reward values $r_i,\,i\in\{1,\dots,N\}$ and the constraint $\eta$ are provided. For each arm, LinConTS assigns independent uniform priors distributed as $\text{Beta}(\alpha_{i,0}=1, \beta_{i,0}=1)$. At every time step $t=1,\dots,T$, LinConTS obtains \emph{Thompson samples} $\widetilde{\mu}_{i,t}\sim \text{Beta}(\alpha_{i,t-1}, \beta_{i,t-1})$ and subsequently solves a LP parameterized by these samples to obtain the instantaneous, probabilistic arm selection vector $x_t=[x_{1,t},\dots,x_{N,t}]$. This vector is sampled to obtain the playing arm, $i(t)\sim x_t$, if the LP is feasible. Subsequently, the reward event $c_{i(t),t}\sim \text{Bern}(\mu_{i(t)})$ is observed and the reward $c_{i(t)}r_{i(t)}$ is collected. The parameter distribution for the played arm is updated according to the rule $\text{Beta}(\alpha_{i(t),t-1}, \beta_{i(t),t-1})\leftarrow (c_{i(t),t}, 1-c_{i(t),t})$.

\subsection{Regret and Violation Upper Bounds}

The theoretical analysis of TS is challenging, and as such only an empirical evaluation of its performance was available until recently~\cite{chapelle2011empirical}. The chief reason for TS' theoretical intractability is that, owing to its randomized nature, novel techniques are required to bound the number of draws for suboptimal arms, which were first introduced in~\cite{agrawal2012analysis} and extended in~\cite{kaufmann2012thompson, agrawal2013further} . At a high level, these techniques compare the Thompson samples for each arm with carefully selected thresholds related to empirical estimates of the arm parameters and their true values. Subsequently, by using known bounds on the sums of Bernoulli random variables, these techniques bound the expected number of times that any suboptimal arm is played.

In case of LinConTS, the selected arm in round $t$ depends on the solution of LP($\widetilde{\boldsymbol\mu}_t$). Therefore, to bound the probability of selecting a suboptimal arm, we need to analyze the LP in~\eqref{eq:linear_prog} with perturbed reward event means. We address this challenge by formulating the Lagrangian dual of the LP that can be solved to obtain specific thresholds for each suboptimal arm, which, when breached under certain well-specified conditions, lead to the suboptimal arm being selected with a nonzero probability. We use these thresholds in combination with the proof technique of~\cite{agrawal2013further} to show that for the suboptimal arms, the expected number of plays number is bounded by $O(\log T)$. A corollary to this result is that the optimal arms are selected at a linear rate. Consequently, the Thompson samples for the optimal arms converge to their true means polynomially fast. As a consequence, the optimal arms are selected with a frequency that does not deviate from the stationary optimal policy by more than $O(\sqrt{T})$. 

Under the assumption of no degenerate cases and that the reward-maximizing arm does not simultaneously satisfy the constraint, exactly two arms support~\eqref{eq:linear_prog}. This well-known result for LPs satisfies the following intuition: the optimal solution to~\eqref{eq:linear_prog} frequently selects a high-reward-value arm, which comes at the cost of frequent failures (i.e., no-reward events). Hence, to satisfy the constraint, the optimal solution sometimes picks another, low-reward-value arm that has a high success probability. Without loss of generality, we denote the optimal arm indices with $1$ and $2$ respectively with parameters that satisfy $0<\mu_1<\eta<\mu_2\leq 1$ and $r_1\mu_1>r_2\mu_2$. Subsequently, we obtain the following regret and violation bounds:
\begin{theorem}
The expected regret for LinConTS,
\begin{align*}
    &\mathbb{E}\big[\mathcal{R}(T)\big]\\
    &\leq\bigg[\sum_{\substack{i\neq\{1, 2\}}}\frac{(1+\gamma)^2}{d(\mu_i,\xi_i)}\Delta_i^+\bigg]\log T+\Delta^+_{\max}\cdot 18\sqrt{2T\log 2}+O(\frac{N}{\gamma^2}),
\end{align*}
where $\xi_i=\frac{(r_1-r_2)\mu_1\mu_2}{(r_i - r_2)\mu_2-(r_i-r_1)\mu_1}>\mu_i$, $\gamma\in(0,1]$ $d(\mu_i,\xi_i)=\mu_i\log\frac{\mu_i}{\xi_i}+(1-\mu_i)\log\frac{1-\mu_i}{1-\xi_i}$, $\Delta_i^+=\max\{0,r^*-\mu_ir_i\}$, and $\Delta^+_{\max}=\max_{i\in [N]}\Delta^+_i$.
\end{theorem}

\begin{theorem}
The expected violation for LinConTS,
\begin{align}
    &\mathbb{E}\big[\mathcal{V}(T)\big]\nonumber\\
    &\leq\bigg[\sum_{i\neq\{1, 2\}}\frac{(1+\gamma)^2}{d(\mu_i,\xi_i)}\delta_i^+\bigg]\log T+\delta^+_{\max}\cdot 18\sqrt{2T\log 2}+O(\frac{N}{\gamma^2}),\nonumber
\end{align}
where $\delta_i^+=\max\{0,\eta-\mu_i\}$ and $\delta^+_{\max}=\max_{i\in [N]}\delta^+_i$
\end{theorem}

\subsection{LP Perturbation Analysis}
\label{sec:lp_pertub}

The Lagrangian dual $\mathcal{L}(\boldsymbol{x},\lambda,\nu,\boldsymbol\psi)$ of the primal LP in~(\ref{eq:linear_prog}) can be formulated as
\begin{align}
    \mathcal{L}(\boldsymbol{x},\lambda,\nu,\boldsymbol\psi) = &-\sum_ix_i\mu_ir_i \nonumber \\
    &+ \lambda\bigg(\eta - \sum_ix_i\mu_i\bigg) \nonumber \\
    &+\nu\bigg(\sum_ix_i - 1\bigg) \nonumber \\
    &-\sum_i\psi_ix_i, \nonumber \\
    = \lambda\eta-\nu &- \sum_i(r_i\mu_i+\lambda\mu_i-\nu+\psi_i)x_i, 
    \label{eq:dual_problem}
\end{align}
where $\boldsymbol{x}\in\mathbf{R}^N$ is the optimization variable, and $\lambda\in\mathbf{R}$, $\nu\in\mathbf{R}$, and $\boldsymbol\psi\in\mathbf{R}^N$ are the Langrange dual variables. The corresponding dual function
\begin{eqnarray}
    g(\lambda,\nu,\boldsymbol\psi) =& \inf_x\,\mathcal{L}(\boldsymbol{x},\lambda,\nu,\boldsymbol\psi)\qquad\qquad\qquad\qquad\qquad \\
      =& \begin{cases}
     \lambda\eta-\nu\quad r_i\mu_i+\lambda\mu_i-\nu+\psi_i = 0,\\
     \qquad\qquad\lambda \geq 0, \\
     \qquad\qquad\psi_i \geq 0.\\
     -\infty\quad\quad\text{otherwise.}
  \end{cases}
    \label{eq:dual_fn}    
\end{eqnarray}
We assume that a strictly feasible solution exists to the $LP(\boldsymbol\mu)$, otherwise there can be no policy that achieves the objective. Then, strong duality holds from Slater's condition and the primal optimal is equal to the dual optimal, i.e., there no duality gap. Hence, from the KKT optimality conditions,
\begin{align}
    \lambda^*\big(\eta - \sum_ix_i^*\mu_i\big)&=0 \\
    \nu^*\big(\sum_ix_i^* - 1\big)&=0,\,\text{and}\\
    \sum_i\psi_i^*x_i^*&=0,
\end{align}
where $\lambda^*,\nu^*,\boldsymbol\psi^*$ are the optimal duals. Since $\psi_i^*\geq 0,x_i^*\geq 0$, each term of $\sum_i\psi_i^*x_i^*$ is non-negative. Consequently for $\sum_i\psi_i^*x_i^*=0$ to hold, $x_i^*>0$ implies that $\psi_i^*=0$ and $x_i^*>0$ implies that $\psi_i^*=0$. Under the assumption of no degenerate cases, strict complementarity holds, i.e., either $x_i^*>0$ or $\psi_i^*>0$~\cite[Chap.~5]{boyd2004convex}.

\emph{Optimal arms:} The arms for which $\psi_i^*=0$ are assigned positive probabilities $x_i^*>0$ by the solution to $LP(\boldsymbol\mu)$. We can then obtain the optimal dual variables by solving the system of linear equations
\begin{align*}
  r_i\mu_i +\lambda^*\mu_i-\nu^* &= 0,\,i\in\{j:\psi_j^*=0,\,j\in\{1,\dots,N\}\},
\end{align*}
which gives
$    \lambda^* = \frac{r_1\mu_1-r_2\mu_2}{\mu_2-\mu_1} \geq 0$ and $ \nu^* = \frac{(r_1-r_2)\mu_1\mu_2}{\mu_2-\mu_1}$.
Note that $\lambda^*$ is the slope of the hyperplane that optimizes $LP(\boldsymbol\mu)$. Further, by solving $\sum_{i=1,2}p_i^*\mu_i=\eta$, we get the optimal selection probabilities $x_1^*=(\mu_2-\eta)/(\mu_2-\mu_1)$ and $x_2^*=(\eta-\mu_1)/(\mu_2-\mu_1)$.

\emph{Suboptimal arms and complementary slackness:} In contrast to the optimal arms, the arms where $x_i^*=0,\,\psi_i^*>0$ constitute suboptimal arms that are assigned zero probability mass by the optimal solution. For these arms, the constraints are \emph{slack} so that sufficiently small perturbations of their parameters does not alter the optimal solution to $LP(\boldsymbol\mu)$. We now quantify the slack for each suboptimal arm $i\not\in\{1,2\}$. For this, we observe that the arms stay suboptimal as long as $\psi_i^*>0$. Rearranging  (\ref{eq:dual_fn}) to obtain $r_i\mu_i +\lambda^*\mu_i-\nu^* =-\psi_i^* < 0$, we get $\mu_i<\frac{\nu^*}{r_i+\lambda^*}$. Hence, we define
\begin{align}
\xi_i := \frac{\nu^*}{r_i+\lambda^*}=\frac{(r_1-r_2)\mu_1\mu_2}{(r_i - r_2)\mu_2-(r_i-r_1)\mu_1}>\mu_i.
    \label{eq:slack}
\end{align}
The value $\xi_i-\mu_i > 0$ is the \emph{complementary slackness} for arm $i$. As long as any (perturbed) mean value, $\mu_i'$, for arm $i$ satisfies $\mu_i' <\xi_i$ the arm $i$ stays suboptimal, i.e, the arm is assigned szero selection probability mass.

\subsection{Arm Selection Bounds}

To bound the regret and violation for LinConTS, we first bound $\mathbb{E}\big[k_i(T+1)\big]$, the expected number of times any suboptimal arm $i\not\in\{1,2\}$ is played until time $T+1$. Our proof technique is inspired by~\cite{agrawal2013further}: First, we show that the probability of playing a sub-optimal arm is a linear function of the probability of playing one of the optimal arms. In order to bound the number of plays of any suboptimal arm, \cite{agrawal2013further} carefully selected \emph{threshold} values that were used to compare the sampled suboptimal and optimal arm means. However, for the constrained problem studied here, the threshold depends on the slack $\xi_i$ for the $i^\text{th}$ suboptimal arm, and the comparison between sampled arm means relies on the LP perturbation analysis above. Next, we show that the coefficient of this linear function decreases exponentially with successive plays of the arm. For this, we use the concentration results from \cite{agrawal2013further} to upper bound the number of plays for each suboptimal arm. We begin by defining the following:

\textbf{Definition 1:} The number of successes observed for arm $i=1,\dots,N$ until time step $t-1$ is denoted by $S_i(t)$. The empirical mean at time $t$ is $\hat{\mu}_i(t)=\frac{S_i(t)}{k_i(t)}$, with $\hat{\mu}_i(t)=1$ when $k_i(t)=0$.

\textbf{Definition 2:} The history of plays until time $t-1$ are denoted by the \emph{filtration} $\mathcal{F}_{t-1}$, i.e.,
\begin{align*}
\mathcal{F}_{t-1} = \{i(w), c_{i(w),w},\,w=1,\dots,t-1\}.    
\end{align*}

\textbf{Definition 3:} For each suboptimal arm $i\not\in\{1,2\}$, we choose two thresholds $y_i$ and $z_i$ such that $\mu_i<y_i<z_i<\xi_i$, which are set at appropriate points of the proof. Further, we define $L_i(T)=\frac{\log T}{d(y_i,z_i)}$. 

\textbf{Definition 4:} We define the variable, $\kappa_1,\dots,\kappa_N$ such that 
$$
\frac{\kappa_jr_j-z_ir_i}{\kappa_j-z_i}=\lambda^*,
$$
where $\kappa_i := z_i$. The hyperplane supported by any two points $(z_i,z_ir_i)$ and $(\kappa_j,\kappa_jr_j),\,j\neq i$ runs parallel to the optimal hyperplane for $LP(\boldsymbol\mu)$.

\textbf{Definition 5:}
We define the variable 
$$
\epsilon_{1,i} = \frac{\kappa_2-\eta}{\kappa_2-\kappa_1}>x_{1,t}>0,
$$
which lower bounds the selection probability mass assigned to arm $1$, $x_{1,t}$,  under certain conditions described later in the proof.

\textbf{Definition 6:} We define two events: $E_i^\mu(t):\{\hat{\mu}_i(t)\leq y_i\}$ and $E_i^\theta(t):\{\theta_{i,t}\leq z_i\}$, where $\theta_{i,t}:=\widetilde{\mu}_{i,t}$ denotes the Thompson sample at time $t$. Intuitively, $E_i^\mu(t)$ and $E_i^\theta(t)$ denote the events that the empirical and sampled means for arm $i$ at time $t$ do not exceed the true arm mean by a large amount.

\textbf{Definition 7:} We define the probability
\begin{align}
    p_{i,t} &= \text{Pr}\big(\{\theta_{1,t}>\kappa_1\}\big|\mathcal{F}_{t-1}\big) 
    \label{eq:arm_1_thresh}
\end{align}


\setcounter{theorem}{0}

Next, we prove the following lemma that establishes the relationship between the number of plays of any suboptimal arm $i\not\in{1,2}$ and the optimal arm $i= 1$.

\begin{lemma}
For all $t\in\{1,\dots,T\}$, and $i\not\in\{1,2\}$, 
\begin{align}
    \text{Pr}&\big(i(t)=i,E_i^\mu(t), E_i^\theta(t)|\mathcal{F}_{t-1}\big)\nonumber \\
    &\leq \frac{1}{\epsilon_{1,i}}\cdot\frac{1-p_{i,t}}{p_ {i,t}}\cdot\text{Pr}\big(i(t)=1,E_i^\mu(t), E_i^\theta(t)|\mathcal{F}_{t-1}\big).
    \label{eq:lemma1}
\end{align}
\end{lemma}
\begin{proof}
Similar to~\cite[Lemma 1]{agrawal2013further}, we note that $E_i^\mu(t)$ is determined by $\mathcal{F}_{t-1}$. Further, when $E_i^\mu(t)$ is false, the left hand side of~\eqref{eq:lemma1} is zero and the inequality is trivially satisfied. Hence, we assume that $\mathcal{F}_{t-1}$ is such that $E_i^\mu(t)$ is true . Subsequently, to prove~\eqref{eq:lemma1}, we show that under a suitably chosen set of conditions $M_i(t)$ that hold with a non-zero probability, the following relationships also hold:
\begin{align}
    \text{Pr}\big(i(t)=i|E_i^\theta(t),\mathcal{F}_{t-1}\big)&\leq (1-p_{i,t})\text{Pr}\big(M_i(t)|E_i^\theta(t),\mathcal{F}_{t-1}\big),\nonumber \\
    \text{Pr}\big(i(t)=1|E_i^\theta(t),\mathcal{F}_{t-1}\big)  &\geq (\epsilon_{1,i}\cdot p_{i,t})\text{Pr}\big(M_i(t)|E_i^\theta(t),\mathcal{F}_{t-1}\big), \nonumber
\end{align}
which immediately gives the Lemma above. The proof details are provided in Appendix A.
\end{proof}

Subsequently, to bound the expected number of plays for any suboptimal arm $i$, we use the following three results available in the literature:~\cite[Lemma 2]{agrawal2013further}
\begin{align}
    &\sum_{t=1}^T\text{Pr}(i(t)=i,\overline{E_i^\mu(t)})\leq \frac{1}{d(y_i,\mu_i)} + 1, \label{eq:agr_lemma2}
\end{align}
which relies on Chernoff-Hoeffding bounds on the concentration of the empirical mean around the true mean,
~\cite[Lemma 3]{agrawal2013further}
\begin{align}
    &\sum_{t=1}^T\text{Pr}(i(t)=i,E_i^\mu(t),\overline{E_i^\theta(t)})\leq L_i(t) + 1, \label{eq:agr_lemma3}
\end{align}
which is based on the fact that after the arm has played a sufficient number of times, $L_i(t)$, the Thompson sample will be close to its mean value, and
~\cite[Lemma 4]{agrawal2013further}
\begin{eqnarray}
    \mathbb{E}\bigg[\frac{1}{p_{i,\tau_j+1}}\bigg]\leq
    \begin{cases}
     1+\frac{3}{\Delta_i'} \qquad\qquad\qquad\quad j<\frac{8}{\Delta'_i},\\
    1+O\big(e^{-\frac{{\Delta'}_i^2j}{2}}\\
    \quad\quad+\frac{1}{(j+1){\Delta'}_i^2}e^{-D_ij}\quad j>\frac{8}{{\Delta'}_i}\\
    \quad\quad+\frac{1}{e^{{\Delta'}_i^2j/4}-1}\big), 
  \end{cases},
    \label{eq:agr_lemma4}
\end{eqnarray}
where $\tau_j$ denotes the time step for the $j^\text{th}$ trial of arm $1$, ${\Delta'}_i=\mu_1-\kappa_1$ and $D_i=z_i\log\frac{z_i}{\mu_1}+(1-z_i)\log\frac{1-z_i}{1-\mu_1}$. This last result uses novel algebraic analysis developed from the concentration of Binomial sums. Based on the above, we obtain the following bound on the expected number of plays of the suboptimal arm $i$ until time $T$:  

\begin{lemma}
 The expected number of plays for any stationary suboptimal arm, $i\in\{1,\dots,K\}\setminus \{1,2\}$, is upper bounded by
 \begin{align}
     \mathbb{E}\big[k_i(T)\big]\leq& 2 + L_i(T) + \frac{1}{d(y_i,\xi_i)} +  \frac{24}{\epsilon_{1,i}{\Delta'_i}^2} \nonumber\\ 
     &+ \frac{1}{\epsilon_{i,i}}\sum_{j=0}^{T-1} O\bigg(e^{-\frac{{\Delta'}_i^2j}{2}} +\frac{1}{(j+1){\Delta'}_i^2}e^{-D_ij}\nonumber\\
    &\qquad\qquad\qquad+\frac{1}{e^{{\Delta'}_i^2j/4}-1}\bigg) .
 \end{align}
\end{lemma}
\begin{proof}
This proof essentially sums over the expressions obtained  in~\eqref{eq:agr_lemma2},~\eqref{eq:agr_lemma3}, and~\eqref{eq:agr_lemma4} until time $T$. The details are provided in Appendix B.
\end{proof}

\begin{lemma}
For the stationary optimal arms $i\in\{1,2\}$,
\begin{align}
    \sum_{i=1,2}\Delta_i\mathbb{E}\big[k_i(T)\big]&\leq \Delta^+_{\max}\cdot 18\sqrt{2T\log 2}, \nonumber\\
    \sum_{i=1,2}(\eta-\mu_i)\mathbb{E}\big[k_i(T)\big]&\leq \delta^+_{\max}\cdot 18\sqrt{2T\log 2},
    \label{eq:reg_vio_optimal}
\end{align}
where $\Delta_i=r^*-\mu_ir_i$, $\Delta^+_{\max}=\max_{i\in [N]}\Delta^+_i$ and $\delta^+_{\max}=\max_{i\in [N]}\delta^+_i$.
\end{lemma}
\begin{proof}
From Lemma 2, the rate of playing any suboptimal is a logarithmic function of $T$. Hence, the set of optimal arms is played at a linear rate. Viewing the play of optimal arms as a two-armed bandit, we can use existing results for constrained TS to bound the regret and violation contribution for the rounds when only the optimal arms are assigned nonzero arm selection probabilities by $LP(\widetilde{\boldsymbol{\mu}}_t)$. First, we decompose the regret and violation expressions separately using the decomposition in~\cite[EC.4]{ferreira2018online} that relies on a frequentist upper and lower bounds that holds with high probability for every $t\in[T]$. Subsequently, we apply~\cite[Lemma EC.2]{ferreira2018online}, which provides an upper bound on the expected regret and violation with respect to the frequentist upper and lower bounds. This directly gives us the bounds in~\eqref{eq:reg_vio_optimal} where we set $K=2$ since only the rounds where the optimal arms are played are taken into account.

\end{proof}

\subsection{Proof of Regret and Violation Bounds}

For some $\gamma\in(0,1]$ we choose the thresholds $y_i\in(\mu_i,\xi_i)$ and $z_i\in(y_i,\xi_i)$ such that $d(y_i,\xi_i)=d(\mu_i,\xi_i)/(1+\gamma)$ and $d(y_i,z_i)=d(y_i,\xi_i)/(1+\gamma)=d(\mu_i,\xi_i)/(1+\gamma)^2$. This leads to
\begin{align*}
    L_i(T) = \frac{\log T}{d(y_i,z_i)} = (1+\gamma)^2\frac{\log T}{d(\mu_i,\xi_i)}.
\end{align*}
Following the ideas in~\cite{agrawal2013further}, we obtain $y_i-\mu_i\geq\frac{\gamma}{1+\gamma}\cdot d(\mu_i,\xi)/\log{\frac{\xi_i(1-\mu_i)}{\mu_i(1-\xi_i)}}$, which gives $\frac{1}{d(y_i,\mu_i)}\leq \frac{2}{(y_i-\mu_i)^2}=O(1/\gamma^2)$.

\textbf{Proof of regret bound:} From Lemma 2 and Lemma 3, we obtain
\begin{align}
    &\mathbb{E}\big[\mathcal{R}(T)\big]=\bigg[\sum_i\Delta_i\mathbb{E}\big[k_i(T)\big]\bigg]_+\nonumber\\
    &\leq\bigg[\sum_{i\neq\{1,2\}}\Delta_i\mathbb{E}\big[k_i(T)\big]\bigg]_+ + \bigg[\sum_{i=1,2}\Delta_i\mathbb{E}\big[k_i(T)\big]\bigg]_+\nonumber\\
    &\leq \sum_{i\neq\{1,2\}}\Delta_i^+\mathbb{E}\big[k_i(T)\big] + \Delta^+_{\max}\cdot 18\sqrt{2T\log 2}\nonumber\\
    &\leq \sum_{i\neq\{1, 2\}}(1+\gamma)^2\frac{\log T}{d(\mu_i,\xi_i)}\Delta_i^++O(\frac{N}{\gamma^2})\nonumber\\
    &\qquad+\Delta^+_{\max}\cdot 18\sqrt{2T\log 2},\nonumber
\end{align}
where the first and second inequalities use $[a+b]_+\leq [a]_++[b]_+$. This completes the regret bound.

\textbf{Proof of violation bound:} From Lemma 2 and Lemma 3, we obtain
\begin{align}
    &\mathbb{E}\big[\mathcal{V}(T)=
                  \bigg[\sum_i(\eta - \mu_i)\mathbb{E}\big[k_i(T)\big]\bigg]_+\nonumber\\
          &\leq 
          \sum_{i\neq\{1, 2\}}\mathbb{E}\big[k_i(T)\big]\big[(\eta - \mu_i)\big]_++ O(\log T)\nonumber\\
          &\leq\sum_{i\neq\{1, 2\}}\frac{(1+\gamma)^2\log T}{d(\mu_i,\xi_i)}\delta_i^++O(\frac{N}{\gamma^2})+\delta^+_{\max}\cdot 18\sqrt{2T\log 2}\nonumber \\
          &=\bigg[\sum_{i\neq\{1, 2\}}\frac{(1+\gamma)^2}{d(\mu_i,\xi_i)}\delta_i^+\bigg]\log T+O(\frac{N}{\gamma^2})\nonumber\\
          &\qquad+\delta^+_{\max}\cdot 18\sqrt{2T\log 2},\nonumber
\end{align}
where the first and second inequalities use $[a+b]_+\leq [a]_++[b]_+$. This completes the violation bound.

\begin{figure*}[t!]
\centering
    \subfigure[Regret]{
    \includegraphics[width=0.23\textwidth]{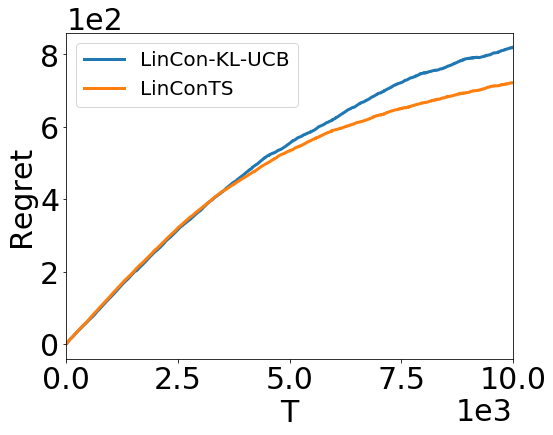}
        \label{fig:reg_coupon}
    }
    \subfigure[Violation]{
    \includegraphics[width=0.23\textwidth]{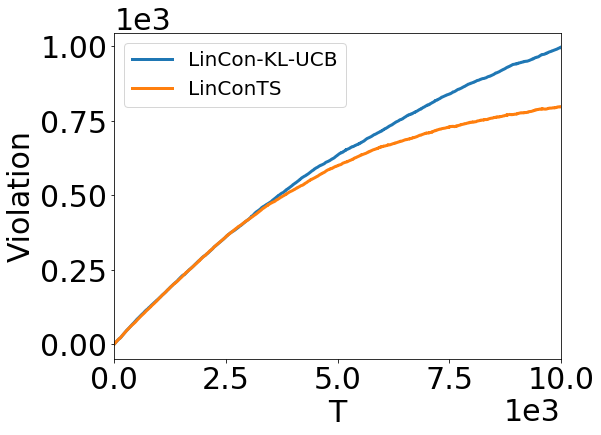}
        \label{fig:vio_coupon}
    }
    \subfigure[Cumulative Reward]{
    \includegraphics[width=0.23\textwidth]{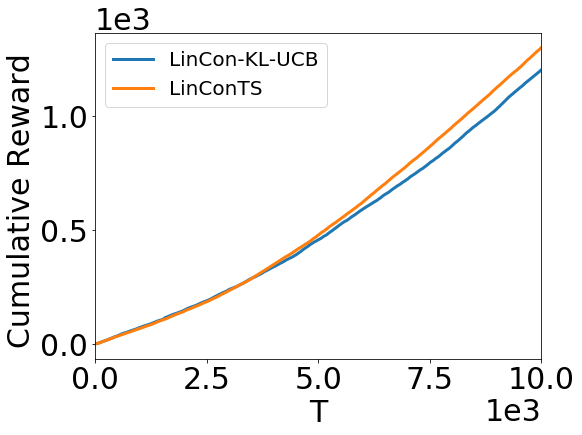}
        \label{fig:rew_coupon}
    }
    \subfigure[Cum. reward / Violation]{
    \includegraphics[width=0.23\textwidth]{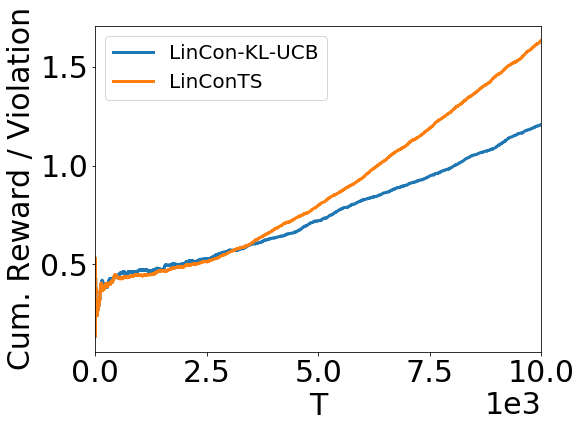}
        \label{fig:rew_vio_coupon}
    }
\caption{ Experimental results for the Coupon-Purchase dataset for $N=142$ coupons with $\eta=0.25$. The LinConTS approach ouperforms the competing LinCon-KL-UCB approach by achieving a lower regret and violation, higher cumulative rewards and a higher ratio of cumulative rewards to violation. }
\label{fig:Coupon}
\end{figure*}
\begin{figure*}[t!]
\centering
    \subfigure[Regret]{
    \includegraphics[width=0.23\textwidth]{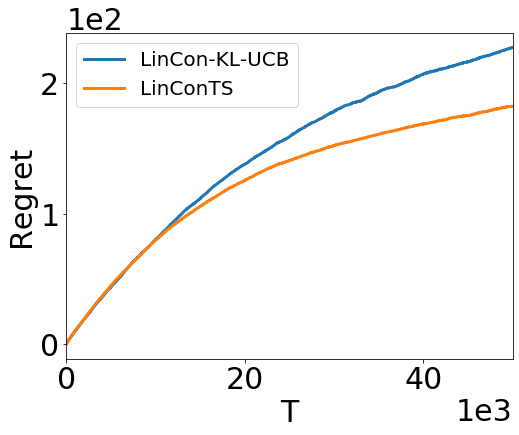}
        \label{fig:reg_edx}
    }
    \subfigure[Violation]{
    \includegraphics[width=0.23\textwidth]{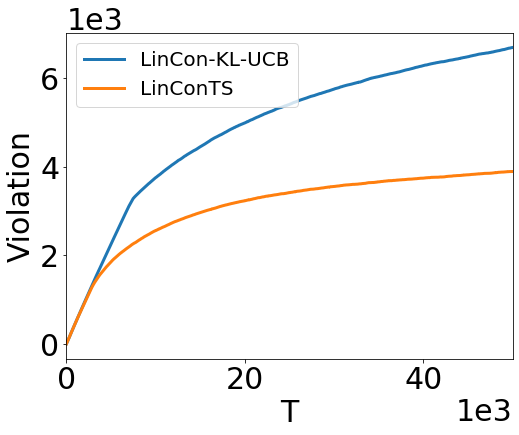}
        \label{fig:vio_edx}
    }
    \subfigure[Cumulative Reward]{
    \includegraphics[width=0.23\textwidth]{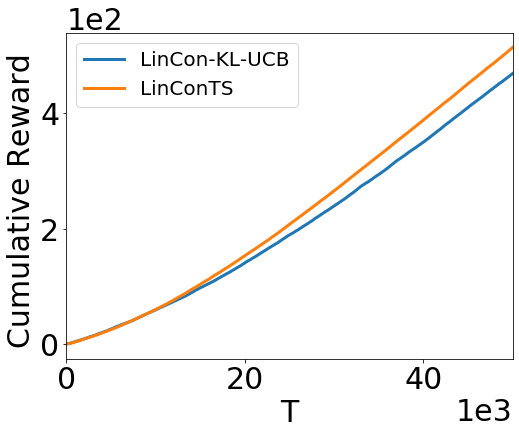}
        \label{fig:rew_edx}
    }
    \subfigure[Cum. reward / Violation]{
    \includegraphics[width=0.23\textwidth]{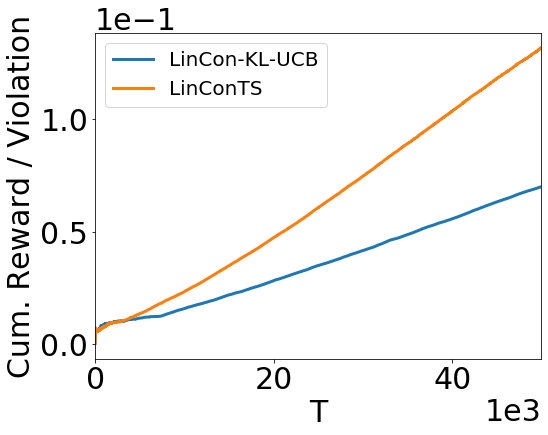}
        \label{fig:rew_vio_edx}
    }
\caption{ Experimental results for the edX-Course dataset for $N=290$ courses with $\eta=0.50$. The LinConTS approach ouperforms the competing LinCon-KL-UCB approach by achieving a lower regret and violation, higher cumulative rewards and a higher ratio of cumulative rewards to violation. }
\label{fig:edx}
\end{figure*}

\section{Related Work}
In~\cite{agarwal2011stochastic}, distribution-free regret bounds for convex optimization with bandit feedback were shown to scale as $O(\sqrt{T})$. Bandits with concave rewards and convex knapsacks were studied within a very general framework in~\cite{agrawal2014bandits}, which subsumed the BwK analysis in~\cite{badanidiyuru2013bandits}. In~\cite{badanidiyuru2013bandits}, an efficient upper confidence bound (UCB)-based approach was introduced that was shown to be optimal for the stochastic BwK problem. Sequential learning under probabilistic constraints has been studied in~\cite{meisami2018sequential}, and Thompson sampling under general problem settings was studied in~\cite{gopalan2014thompson}. In~\cite{xia2015thompson}, a Thompson sampling algorithm for budgeted MABs was proposed that outperforms the UCB BwK algorithm. Subsequently TS was applied to network revenue management in~\cite{ferreira2018online}, were distribution-free bounds that scale as $O(\sqrt{T})$ were shown to hold. In~\cite{chen2018beyond}, a horizon-dependent UCB approach was proposed whose regret and violation performance was shown to scale with $O(\sqrt{T})$. Recently, applications of linearly constrained MABs have been proposed for advertiser portfolio optimization~\cite{pani2017large}, wireless communications~\cite{saxena2019constrained}, and real-time electricity pricing~\cite{tucker2019constrained}. 

\section{Numerical Experiments}
\label{sec:numrel}
We evaluate the regret and violation performance of LinConTS on two real-world datasets, \emph{Coupon-Purchase}~\cite{kaggle2016coupon} and \emph{edX-Course}~\cite{chuang2016harvardx}, respectively. The Coupon-Purchse and edX-Course datasets have been explored previously for constrained bandit problems, albeit in a multi-play setting that allowed multiple arms to be played in each time step and the assumed that the reward values were stochastic~\cite{chen2018beyond, cai2017multi}. The experiments are implemented in Python using Jupyter notebooks and have been made publicly available at~\cite{vidit2019linconts}.

For the Coupon-Purchase dataset, which contains discount coupons applied to online purchases, we extract all the $N=142$ coupons for products priced equal to or below below $200$ price units that have been purchased by at least one customer. For these coupons, we obtain the purchase rate and the discounted selling price from the dataset. The edX-Course dataset contains enrolment information for $N=290$ Harvard and MIT courses. We process this dataset according to previous experiments~\cite{chen2018beyond}: the course participation rates are estimated by max-min normalization of the number of participants in each course, and the course certification rates are obtained by dividing the number of certified participants in each course by the number of course participants. 

We model each coupon in the Coupon-Purchase dataset and each course in the edX-Course dataset using independent bandit arms. Further, for the Coupon-Purchase dataset, we generate independent Bernoulli-distributed reward events with mean values obtained from the purchase rates and a deterministic reward value defined as the final selling price normalized by $200$. For the processed Coupon-Purchase dataset, the reward event means and the rewards values are found to lie between $[0, 0.30]$ and $(0,1]$ respectively. Analogously for the edX-Course dataset, the reward event means are obtained from the course participation rates and the corresponding reward values are the course certification rates that are assumed to be known in advance, e.g. from historical course data. For this dataset, the reward event means and the rewards values are found to span $[0, 1]$ and $(0,0.40]$ respectively.

We implement two linearly constrained bandit algorithms, LinConTS and LinCon-KL-UCB respectively. The LinConTS algorithm is described in Sec.~\ref{sec:linconts} and the pseudocode is available in Alg. 1. The LinCon-KL-UCB, described in Appendix C, is inspired by the ConTS algorithm proposed for multi-play linearly constrained MABs in~\cite{chen2018beyond}. However, compared to ConTS that relies on an index-based UCB, LinCon-KL-UCB estimates the UCB for each arm using the Kullback-Liebler (KL) divergence metric. For the Bernoulli-distributed rewards considered in here, KL-based UCB has been shown to achieve optimal regret and significantly outperform index-based UCB schemes~\cite{garivier2011kl}.

The performance of LinCon-KL-UCB and LinConTS for the Coupon-Purchase dataset and for $\eta=0.25$ is shown in Fig.~\ref{fig:Coupon}, where the results have been averaged over $16$ test runs. LinConTS achieves a lower regret compared to LinCon-KL-UCB in Fig.~\ref{fig:reg_coupon}. This demonstrates that compared to LinCon-KL-UCB, LinConTS is closer to the cumulative rewards achieves by the stationary optimal policy. We study the cumulative violations for each approach in Fig.~\ref{fig:vio_coupon}, where LinConTS demonstrates fewer constraint violations than LinCon-KL-UCB. Interestingly, in Fig.~\ref{fig:rew_coupon}, LinConTS achieves higher cumulative rewards as well compared to LinCon-KL-UCB. We combine the related effects of regret and violation minimization by calculating the ratio between the cumulative rewards and the cumulative violations at each time step. This quantity, depicted in Fig.~\ref{fig:rew_vio_coupon}, can be interpreted as the additional reward earned for every constraint violation. We observe that LinConTS achieves a higher ratio, which demonstrated that LinConTS is more efficient in exploiting the infrequent constraint violations.

For the edX-Course dataset and $\eta=0.50$, the experimentalt results for LinConTS and LinCon-KL-UCB schemes are averaged over $16$ test runs and shown in Fig.~\ref{fig:edx}. Similar to the results for the Coupon-Purchase dataset, here also LinConTS has lower regret and violation than LinCon-KL-UCB in Fig.~\ref{fig:reg_edx} and Fig.~\ref{fig:vio_edx} respectively. Also, from Fig.~\ref{fig:rew_edx} and Fig.~\ref{fig:rew_vio_edx}, LinConTS increases the cumulative reward, and the ratio of cumulative reward to violations, respectively.

\section{Conclusions and Further Work}

For constrained bandit problems, LP subroutines enable powerful sequential arm selection techniques. Combined with an underlying Thompson Sampling approach, these techniques promise efficient and robust solutions that are optimal in terms of their learning rate. We have addressed the MAB problem of maximizing the cumulative reward when the reward event probability is constrained above a fixed threshold in every round. For this problem, we described LinConTS, which incorporates a LP subroutine in every step of the Thompson Sampling heuristic. We have provided the first instance-dependent finite-time analysis our algorithm. Through numerical results for two real-world datasets, we have showed that LinConTS outperforms an optimal UCB-based algorithm in terms of the regret and violation metrics. 

We have considered a specific type of linear constraint in this paper. However, the proof technique developed in this paper can inspire solutions to other constrained MAB problems, for example to problems that deal with more general constraints. Further, it could be useful to develop a composite notion of regret and develop optimal lower bounds, for example by leveraging the recent results in~\cite{garivier2019explore}.

\section*{Acknowledgements}

This work was partially supported by the Wallenberg Artificial Intelligence, Autonomous Systems and Software Program (WASP) funded by Knut and Alice Wallenberg Foundation. We thank Simon Lindst\r{a}hl for noting a flaw in the original formulation of Lemma~3, which has been updated in this version of the paper.

\bibliography{refs}
\bibliographystyle{apalike}

\section*{Appendix A}

Consider the Thompson samples $\theta_{i,t},\,i=1,\dots,N$ for any round indexed by $t$. The samples for stationary optimal arm $1$ and stationary suboptimal arm $i$ are denoted by $\theta_{1,t}$ and $\theta_{i,t}$ respectively. Let $M_i(t)$ denote the event
\begin{eqnarray}
    M_i(t):\,\begin{cases}\theta_{j,t}\leq \kappa_j,\,j\not\in\{1,2,i\}\,,\text{ and}\\
    \kappa_2 <\theta_{2,t}\leq \mu_2,
    \end{cases}
\end{eqnarray}
where we choose thresholds such that $\kappa_2>\eta$. Consider the probability of the event where suboptimal arm $i$ is selected under the filtration $\mathcal{F}_{t-1}$ and the Thompson sample $\theta_{i,t}$ such that $E_i^\theta$ is true, i.e., $\text{Pr}(i(t)=i|E_i^\theta(t),\mathcal{F}_{t-1})$. We have that arm $i$ is a part of the optimal solution only if arm $1$ is below the threshold $\kappa_1$, all other stationary arms are below their respective thresholds $\kappa_j,\,j\not\in\{1,2,i\}$, and the stationary optimal arm $2$ is above $\kappa_2$ (in which case at least some of the optimal solutions are supported by arms $i$ and $2$). Hence, we have
\begin{align}
    \text{Pr}&(i(t)=i|E_i^\theta(t),\mathcal{F}_{t-1})\nonumber \\
    &\leq \text{Pr}(i(t)=i, \theta_1(t)<\kappa_1,M_i(t)|E_i^\theta(t),\mathcal{F}_{t-1})\nonumber\\
    &= (1-p_{i,t})\text{Pr}(M_i(t)|E_i^\theta(t),\mathcal{F}_{t-1}),
    \label{eq:lemma1_subopt}
\end{align}
where the second step follows from the independence of events conditional on the filtration $\mathcal{F}_{t-1}$.

Next, we bound the probability of selecting arm $1$. We observe that, conditioned on $M_i(t)$ and $E_i^\theta$, arm $1$ forms a part of the optimal solution at time $t$ along with arm $2$. Further, the probability mass assigned to arm $1$ is $(\theta_{2,t}-\eta)/(\theta_{2,t}-\theta_{1,t})$. For any Thompson samples such that $\theta_2>\kappa_2$ and $\theta_1>\kappa_1$, the probability mass assigned to arm $1$ is at least $(\kappa_2-\eta)/(\kappa_2-\kappa_1)=\epsilon_{1,i}$. Consequently, we have
\begin{align}
    &\text{Pr}\big(i(t)=1|E_i^\theta(t),\mathcal{F}_{t-1}\big) \nonumber \\
    &\geq \text{Pr}\big(i(t)=1,M_i(t)|E_i^\theta(t),\mathcal{F}_{t-1}\big) \nonumber \\
    &= \text{Pr}\big(M_i(t)|E_i^\theta(t),\mathcal{F}_{t-1}\big)\cdot\text{Pr}\big(i(t)=1|M_i(t),E_i^\theta(t),\mathcal{F}_{t-1}\big) \nonumber \\
    &\geq \epsilon_{1,i}\cdot p_{i,t}\cdot\text{Pr}\big(M_i(t)|E_i^\theta(t),\mathcal{F}_{t-1}\big),
    \label{eq:lemma1_opt}
\end{align}

Combining~\eqref{eq:lemma1_subopt} and~\eqref{eq:lemma1_opt} we get the desired result.

\newpage

\section*{Appendix B}

Similar to the approach in~\cite{agrawal2013further}, we bound the number of plays of any suboptimal arm in the following manner:
\begin{align*}
    \mathbb{E}\big[k_i(T)\big] =&  \sum_{t=1}^T\text{Pr}(i(t)=i)  \\
    =& \sum_{t=1}^T\text{Pr}(i(t)=i,E_i^\mu(t),E_i^\theta(t)) \\
    &+ \sum_{t=1}^T\text{Pr}(i(t)=i,E_i^\mu(t),\overline{E_i^\theta(t)}) \\ &+\sum_{t=1}^T\text{Pr}(i(t)=i,\overline{E_i^\mu(t)}) \\
\end{align*}
The last two terms of this expression are upper bounded by (17) and (16) respectively. Then, following the approach in \cite{agrawal2013further}, we bound the first term of the expression above using Lemma 1, where we exploit the fact that the number of plays of arm $i$ are a linear function of the number of playes of arm $1$,
\begin{align*}
    \sum_{t=1}^T&\text{Pr}(i(t)=i,E_i^\mu(t),E_i^\theta(t)) \\
    &=\sum_{t=1}^T\mathbb{E}\bigg[\text{Pr}(i(t)=i,E_i^\mu(t),E_i^\theta(t)\big|\mathcal{F}_{t-1})\bigg]\\ 
    &\leq \sum_{t=1}^T\mathbb{E}\bigg[\frac{1-p_{i,t}}{\epsilon_{1,i}\cdot p_{i,t}}\text{Pr}(i(t)=1,E_i^\mu(t),E_i^\theta(t)\big|\mathcal{F}_{t-1})\bigg] \\
    &= \sum_{t=1}^T\mathbb{E}\bigg[\mathbb{E}\bigg[\frac{1-p_{i,t}}{\epsilon_{1,i}\cdot p_{i,t}}I(i(t)=1,E_i^\mu(t),E_i^\theta(t)\big)\bigg]\bigg] \\
    &\leq\sum_{k=0}^{T-1}\mathbb{E}\bigg[\bigg(\frac{1}{\epsilon_{1,i}\cdot p_{i,\tau_k+1}} - 1\bigg)\sum_{t=\tau_k+1}^{\tau_{k+1}}I(i(t)=1)\bigg] \\
    &=\sum_{k=0}^{T-1}\mathbb{E}\bigg[\frac{1}{\epsilon_{1,i}\cdot p_{i,\tau_k+1}} - 1\bigg],
\end{align*}
where $I$ is the indicator function, and we have used the fact that $\epsilon_{1,i}$ is independent of the history of plays. From (18), we have an upper bound on $\mathbb{E}(\frac{1}{p_{i,\tau_j+1}})$. By collecting the upper bounds from (16), (17), and (18), we directly obtain Lemma 2.

\newpage

\section*{Appendix C}

We present the pseudocode for the LinCon-KL-UCB algorithm in this section.

\begin{algorithm}
   \caption{LinCon-KL-UCB}
   \label{alg:lin_con_kl_ucb}
\begin{algorithmic}[1]
   \STATE {\bfseries Input:} Reward Values $r_{\{1,\dots,N\}}$, Constraint $\eta$, $c$
   \STATE \textbf{Initialize:} $k_{\{1,\dots,N\},0}=0,s_{\{1,\dots,N\},0}=0$.
   \FOR{ Time index $t=1$ {\bfseries to} $T$ }
   \IF{ $t<N$ }
   \STATE $i(t) = t$
   \ELSE
   \FOR{ Arm index $i=1$ {\bfseries to} $N$ }
   \item $\tilde{\mu}_{i,t} = \max\big\{q\in\Theta:k_id\big(\frac{k_i}{s_i},q\big)\log(t)+c\log\log(t)\big\}$
   \ENDFOR
   \STATE Solve, if feasible, the linear program:
   \begin{align*}
  LP(\tilde{\mu}_t):\,\text{maximize}\,\sum_i x_{i,t}\tilde{\mu}_{i,t}r_i \qquad\qquad\qquad\qquad
  \end{align*}
  \begin{equation}
  \qquad\text{ subject to } \begin{cases}
     \sum_i x_{i,t}\tilde{\mu}_{i,t} \geq \eta\\
    \sum_ix_{i,t} = 1\\
    x_{i,t} \geq 0\quad \forall\,i\in\{1,\dots,N\}
  \end{cases},
  \label{eq:linconts}
\end{equation}
   \IF{ a (feasible) optimal solution existed } 
    \STATE Sample $i(t)~\sim [x_{1,t},\dots,x_{N,t}]$
   \ELSE  
    \STATE Sample $i(t)$ uniformly from $\{1,\dots,N\}$.
   \ENDIF
   \ENDIF
   \STATE \textbf{Observe:} Reward event $c_{i(t)}\in\{0,1\}$.
   \STATE \textbf{Update:} \\
             $\qquad k_{i(t)} = k_{i(t)} + 1$ \\
             $\qquad s_{i(t)} = s_{i(t)} + c_{i(t)}$.
   \ENDFOR
\end{algorithmic}
\label{alg:linconts}
\end{algorithm}

\end{document}